\documentclass[sigconf]{acmart}

\usepackage[utf8]{inputenc}
\usepackage{amsmath}
\usepackage{amsfonts}
\usepackage{amssymb}
\usepackage{graphicx}
\usepackage{bm} 
\usepackage{subfig}

\usepackage{listings} 
\usepackage{color}    
\definecolor{mygreen}{rgb}{0,0.6,0}
\definecolor{mygray}{rgb}{0.5,0.5,0.5}
\definecolor{mymauve}{rgb}{0.58,0,0.82}

\makeatletter

\hypersetup{
pdfauthor={\@author},
pdftitle={\@title}
}

\begin{document}

\title{calibDB: enabling web based computer vision through on-the-fly camera calibration}

\author{Pavel Rojtberg}
\email{pavel.rojtberg@igd.fraunhofer.de}
\affiliation{%
  \institution{Fraunhofer IGD}}

\author{Felix Gorschlüter}
\email{felix.gorschlueter@igd.fraunhofer.de}
\affiliation{%
  \institution{Fraunhofer IGD}}


\lstset{
commentstyle=\color{mygreen},
stringstyle=\color{mymauve},
showspaces=false,
captionpos=b,
showstringspaces=false,
tabsize=2,
frame=none,
basicstyle=\footnotesize,
caption=\relax
}

\copyrightyear{2019}
\acmYear{2019}
\setcopyright{acmlicensed}
\acmConference[Web3D '19]{Web3D '19: The 24th International Conference on 3D Web Technology}{July 26--28, 2019}{Los Angeles, CA, USA}
\acmBooktitle{Web3D '19: The 24th International Conference on 3D Web Technology (Web3D '19), July 26--28, 2019, Los Angeles, CA, USA}
\acmPrice{15.00}
\acmDOI{10.1145/3329714.3338132}
\acmISBN{978-1-4503-6798-1/19/07}

\begin{abstract}
For many computer vision applications, the availability of camera calibration data is crucial as overall quality heavily depends on it.
While calibration data is available on some devices through Augmented Reality (AR) frameworks like ARCore and ARKit, for most cameras this information is not available.
Therefore, we propose a web based calibration service that not only aggregates calibration data, but also allows calibrating new cameras on-the-fly.
We build upon a novel camera calibration framework that enables even novice users to perform a precise camera calibration in about 2 minutes.
This allows general deployment of computer vision algorithms on the web, which was previously not possible due to lack of calibration data.
\end{abstract}

%
%



\begin{teaserfigure}
  \centering
\subfloat[Calibrated camera matrix only] {
\includegraphics[width=0.32\textwidth]{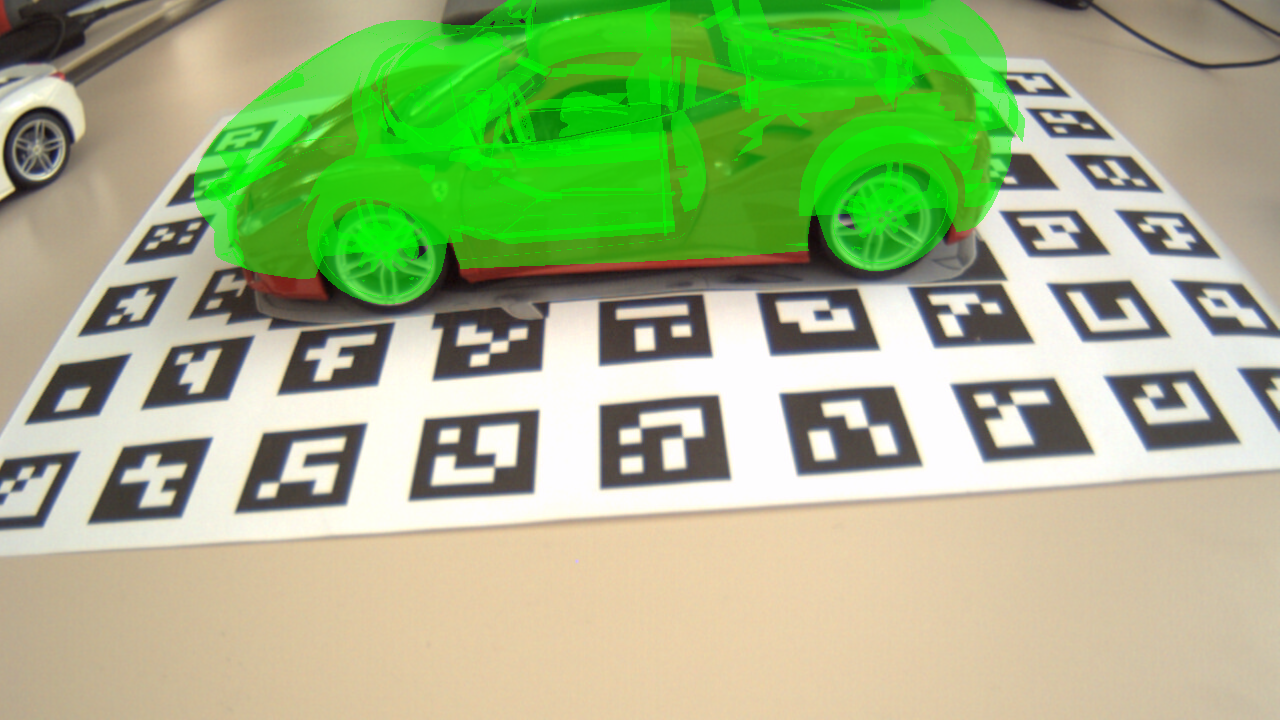}
\label{fig:konly}
}
\subfloat[Rectified image] {
\includegraphics[width=0.32\textwidth]{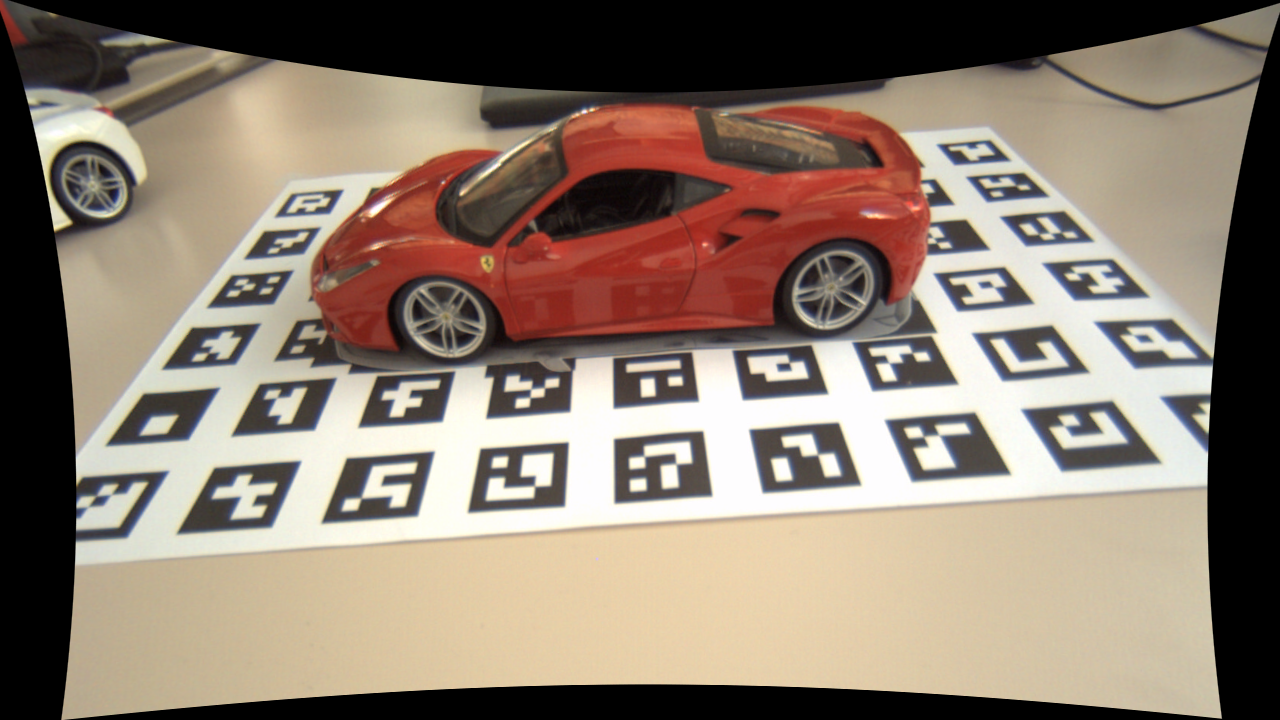}
\label{fig:remap}
}
\subfloat[Fully calibrated camera] {
\includegraphics[width=0.32\textwidth]{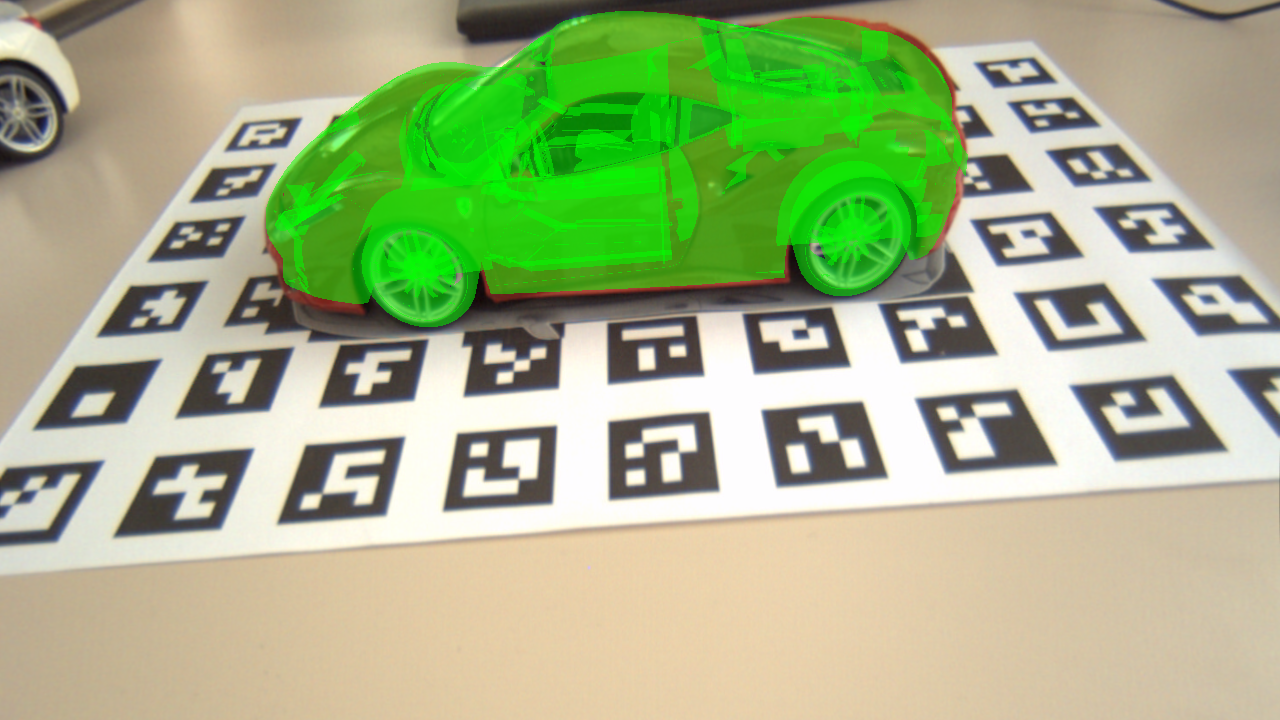}
\label{fig:undistorted}
}
  \caption{Effect of camera calibration on an augmented reality scene: Although a calibrated camera matrix is used in (a), the misalignment is clearly visible. Using a complete distortion model allows rectifying the image (b). Together with an adapted camera matrix, this results in a fully aligned augmentation (c).}
  \label{fig:teaser}
\end{teaserfigure}

\maketitle


\section{Introduction}

Camera calibration in the context of computer vision is the process of determining the internal geometrical and optical camera characteristics (intrinsic parameters) and optionally the position and orientation of the camera frame in the world coordinate system (extrinsic parameters).
The performance of many 3D vision algorithms directly depends on the quality of this calibration \cite{furukawa2008accurate}.
Furthermore, calibration is a recurring task that has to be performed each time the camera setup is changed. Even cameras of the same series can have different intrinsic parameters due to build inaccuracies.

Native applications can leverage frameworks like ARKit and ARCore which provide the camera intrinsic parameters per-frame. Alternatively developers use lower-level vision libraries like OpenCV \cite{opencv_library} and manually acquire and ship the calibration data specific to their setup.

For web-based computer vision solutions the WebXR Device API Draft \cite{webxr} provides the intrinsic camera matrix through the \textit{XRView} interface. However, the data is encoded into a \textit{projectionMatrix} as used for rendering and needs special conversion to be used with vision algorithms. The lens distortion coefficients are completely absent, which drastically reduces precision (see Figure \ref{fig:teaser}). These two aspects show that the existing API focuses on a camera representation primarily suited for rendering --- likely due to its strong heritage from the WebVR API.
Furthermore, the available WebXR polyfills either leverage ARKit\footnote{\url{https://github.com/mozilla-mobile/webxr-ios}} or ARCore\footnote{\url{https://github.com/googlecodelabs/ar-with-webxr}} to retrieve calibration information thus limiting computer vision applications to these platforms.

Web-based augmented reality (AR) applications using low-level computer vision primitives \cite{gottl2018efficient} are therefore forced to assume a default camera intrinsic, which is imprecise or ship a set of manually acquired calibrations with the aforementioned drawbacks.

Our work therefore aims at providing a camera calibration database that web applications can use to retrieve precise calibration data on-the-fly. The database is designed to be extendable both in terms of calibration models and new cameras. For this we leverage the novel camera calibration framework by \cite{rojtberg2018, rojtberg2019hci} to guide end-users through the calibration process if their camera is not yet included in the database. This enables developers to deploy computer-vision applications to the full diversity of the web platform.

This paper is structured as follows: in Section \ref{sec:background} we introduce the prevalent camera calibration methods and models. In Section \ref{sec:webposecalib} we present our architecture for interactive calibration acquisition on the web. Here, we present our calibration storage and on-demand retrieval as well as proposing necessary extensions to WebXR.
We conclude with Section \ref{sec:conclusion} giving a summary of our results and discussing the limitations and future work.

\section{Background}
\label{sec:background}
In this section we first introduce the computer vision camera terminology as well as common distortion models, that are supported by our calibration service.
Then we turn to current state-of-the-art methods for camera calibration and user guidance.

\subsection{Intrinsic Parameters}
The intrinsic camera parameters that are recovered during calibration are typically the focal length and the principal point, encoded in the camera matrix $\mathbf{K} \in \mathbb{R}^{3x3}$ and a set of lens distortion coefficients $\mathbf{d} = [k_0, \ldots, k_n]$ \cite{hartley2005multiple}.

We can now formalize the mapping of a 3D point in camera space $ \mathbf{P} = [X, Y, Z] $ to a 2D image point $\mathbf{p} = [x, y]$ as
\begin{equation}
\pi \left( \mathbf{P}; \mathbf{d} \right) = \mathbf{K} \, \Delta(\frac{1}{Z} \mathbf{P}).
\label{eq:cvcam}
\end{equation}
Here $\Delta(\cdot)$ is the lens distortion function parameterized by $\mathbf{d}$ and typically models the radial distortion as
\begin{equation}
\Delta_R(\mathbf{p}) =\;\mathbf{p} \left( 1 + k_1 r^2 + k_2 r^4 + k_3 r^6 \right)
\label{eq:dr}
\end{equation}
where $ r = \sqrt{x^2 + y^2}$.

Lens distortion is currently not handled by the WebXR API, which only exposes the camera matrix $\mathbf{K}$. While the effect of lens distortion can be neglected on simple webcams which resemble the pinhole optics, this does not hold generally.

Figure \ref{fig:konly} shows an image captured with the Computar E3Z4518CS lens with an AR-overlay rendered considering $\mathbf{K}$ only. As can be seen the AR-overlay diverges from the image towards the image edges. Rectifying the image by inverting eq. \eqref{eq:dr} and adapting $\mathbf{K}$ accordingly, we can make the overlay fit the image as can be seen in Figure \ref{fig:undistorted}.

Web-based computer vision should not be restricted to webcam imagery, therefore we have to expect all kinds of cameras. Eq. \eqref{eq:dr} is also specified in the DNG image format \cite{dngspec} as \textit{WarpRectilinear} for processing images from interchangeable-lens cameras.

Additionally, the DNG format includes a specialized distortion model for fisheye lenses \cite{kannala2006generic}, \textit{WarpFisheye}:
\begin{equation}
\Delta_F(\mathbf{p}) =\;\mathbf{p} \frac{1}{r} \left( \theta + k_1 \theta^3 + k_2 \theta^5 + k_3 \theta^7 \right)
\end{equation}
where $\theta$ is the angle between the principal axis and an incoming ray.
This model is required as fisheye lenses can expose a field of view $\geq 180^{\circ}$ which cannot be represented using a rectilinear projection.

The OpenCV library supports both $\Delta_R$ and $\Delta_F$ as well as more sophisticated models for e.g. spherical $360^\circ$ cameras \cite{geyer2000unifying} as employed by the street-view cars or spherical video.

To accommodate for the different calibration models our database therefore not only stores the distortion coefficients $\mathbf{d}$, but the full calibration data to be able to fit a new camera model on demand --- without requiring a user to capture new calibration data.

\subsection{Guided camera calibration}
\begin{figure}
\subfloat[ChArUco Pattern] {
\includegraphics[width=0.24\textwidth]{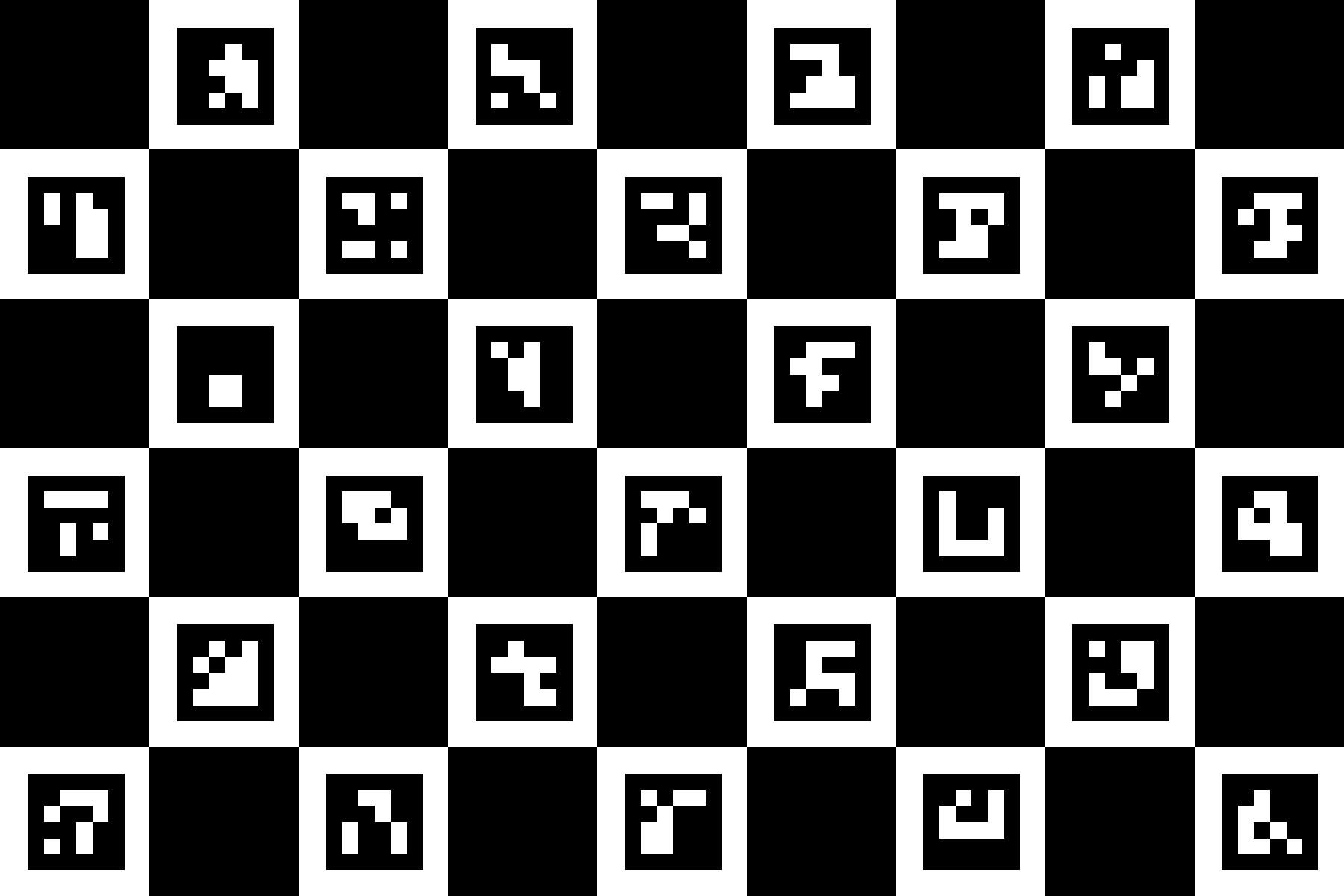}
\label{fig:pattern}
}
\subfloat[Calibration Overlay] {
\includegraphics[width=0.2\textwidth]{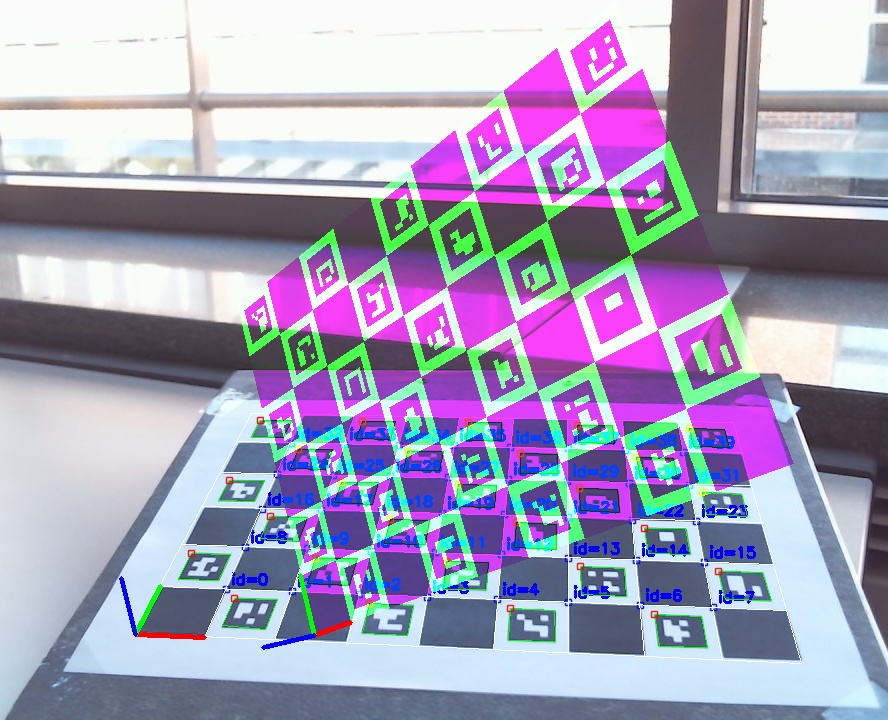}
\label{fig:overlay}
}
\caption{The interface of our calibration guidance}
\end{figure}

The prevalent approach to camera calibration is based on acquiring multiple images of a planar pattern of known size \cite{zhang2000flexible}. These patterns are easy to obtain at high precision using conventional printers or by simply displaying them on a monitor.
Typically, chessboard patterns are used as the chessboard corners provide 2D measurements at sub-pixel precision. However, chessboard detection involves the time-consuming step of ordering the detected rectangles to a canonical grid, which slows down the method below interactive rates.

Therefore, our method uses the ChArUco Pattern (see Figure \ref{fig:pattern}) which interleaves ArUco Markers \cite{garrido2014automatic} within the chessboard. These markers are fast to detect and allow deducing position and orientation of the whole board. Notably, they also allow only a part of the board to be visible.

To acquire calibration data, we build upon the novel camera calibration framework by \cite{rojtberg2018, rojtberg2019hci} that dynamically generates target poses to determine the intrinsic parameters. This way only around 10 images are required to perform a precise calibration. Additionally, this allows displaying an overlay (see Figure \ref{fig:overlay}) to guide to specific poses. The whole process of capturing the images and computing a new calibration on average only requires 2 minutes --- even if the user is not familiar with computer vision.

\section{Web based implementation}
\label{sec:webposecalib}
In this section we describe our calibration service "calibDB" in detail. First we discuss the high-level architecture and internal protocol of the service. Then we describe the external API and data format used for calibration data retrieval and acquisition. Finally, we discuss how the current WebXR API should be extended to seamlessly provide calibration data to computer vision applications.

\subsection{Efficient Client/ Server separation}
To bring our existing OpenCV based implementation to the Web, we utilize the OpenCV.js bindings, that wrap the C++ code with Emscripten \cite{zakai2011emscripten} into a WebAssembly library.
Here, we do not fully port our existing code to javascript to be executed in the browser. Instead, we introduce a client/server split as the captured 2D measurements, and the final calibration parameters will be transferred to the server anyway.
Our architecture is split as follows:
\begin{itemize}
\item A web-based acquisition client, that captures video using WebRTC \cite{burnett2011getusermedia} and performs low-level image processing directly on the device. This reduces latency and offloads the computation heavy image processing from the server.
\item The calibDB server component that receives the captured key-points and provides new target poses to the clients. This allows re-using most of our control logic and keeps the architecture extendable for multiple clients, as is useful with e.g stereo camera calibration.
\end{itemize}

\begin{figure}
\includegraphics[width=0.4\textwidth]{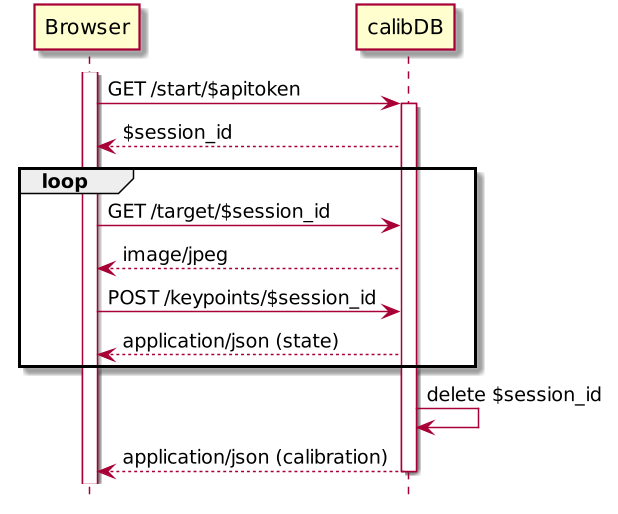}
\caption{The REST protocol of our web-based camera calibration system}
\label{fig:sequence}
\end{figure}

Figure \ref{fig:sequence} shows a sequence diagram of the REST based communication between browser and calibDB. As we want to provide our calibration service publicly on the internet we employ API tokens to prevent abuse. After the client was authorized by calibDB, a session ID is returned that is used to track the calibration session and for further authentication.
The client then asks for a new target pose which is returned as a jpeg image that is composited with the video stream using the "color" blend mode. Our underlying method compares the projected pattern images to check whether the user is sufficiently close to the target pose, therefore we can just use the non-black pixels of the overlay image to extract this information.
Once the target pose was reached the client sends the acquired 2D keypoint positions to calibDB, which returns a JSON-message \cite{bray2017javascript} containing the calibration results or a state indicating that further measurements are needed.

Our client was tested with Google Chrome and Mozilla Firefox. Here, Chrome is preferable as it also provides the USB-ID of the device, which allows differentiating devices of one series that use different hardware (same name, but different sensor).

\subsection{Calibration Database}
\label{sec:calibdb}
The service can be queried for calibration data using a combination of \textit{userAgent}, \textit{MediaStreamTrack} and \textit{MediaTrackSettings} \cite{mediastream} as the key:

\begin{lstlisting}[label=lst:request,language=java, caption=Example calibration-data request]
{
    "camera": "C922 Pro Stream Webcam (046d:085c)",
    "platform": "X11; Linux x86_64",  
    "img_size": [1280, 720],
    "zoom": 0
}
\end{lstlisting}
Here the \textit{camera} property is used for differentiating multiple cameras attached to the PC or the front and back camera on mobile devices. The \textit{host} property is mainly used to differentiate mobile devices where \textit{camera} would only contain "front" or "back". The "zoom" property translates to the currently set focal length of the camera or zero if the focal length cannot be determined.

If no reliable calibration data is available the server responds with the HTTP/307 status code, redirecting to the calibration-guidance landing page as described in Section \ref{sec:background}.

To verify whether calibration data is reliable, we collect at least 5 different calibrations and compute the variance of the intrinsic parameters. Only if the variance is small compared to the parameter values, we consider the calibration data reliable. Here, we aim to enforce re-calibration for interchangeable lens cameras. These identify using the same name, but have largely varying intrinsic properties.
Notably, this also covers the use of manually operated lenses where the "zoom" property cannot be read automatically.

If reliable calibration data is available it is returned in JSON encoding as:

\begin{lstlisting}[label=lst:calib,language=java, caption=Example calibration-data response]
{
    "img_size": [1280, 720],
    "camera_matrix": [[1.43e+03, 0.0, 9.52e+02],
                      [0.0, 1.43e+03, 5.05e+02], 
                      [0.0, 0.0, 1.0]],
    "distortion_coefficients": [ ... ],
    "distortion_model": "rectilinear",
    "avg_reprojection_error": 0.72
}
\end{lstlisting}
The message contains the parameters $\mathbf{K}$ and $\mathbf{d}$ as discussed in Section \ref{sec:background}. Additionally, it provides the resolution at which the calibrated was performed. This is useful when the exact requested resolution is not available. In this case the calibration for closest resolution is returned. The client is now able to either adapt the capturing or redirect to the guidance page, if a specific resolution is crucial.

The client is also able to explicitly specify the desired \textit{distortion\_model}, by adding it to the request (Listing \ref{lst:request}), if only a specific model is supported. In case no calibration using the requested model is available for the specified camera, the server can transparently perform a new parameter fitting on-the-fly.
This is made possible by storing the 2D key-points alongside the calibration results. For instance if $\Delta_R$ is requested, but only calibrations for $\Delta_F$ are available, the server can repeat the parameter fitting using the existing data.
However, this is not always valid. In the example above the \textit{rectilinear} model is not capable of explaining all measurements as produced by a fisheye lens.
Therefore, the response also includes the \textit{avg\_reprojection\_error}, which is the residual error on the measurements.
The client is now again able to redirect to the guidance page to force a more precise calibration.

Our prototype implementation supports the "rectilinear" and "fisheye" distortion models and stores the calibration results as well as the key-points in a schema-less database \cite{mongodb2016mongodb}.
This allows to easily extend the system to new distortion models as needed.

\subsection{Extending the WebXR API}
\label{sec:webxrext}
To provide the relevant calibration information through the WebXR API, it needs to be extended in several ways. We propose to extend the \textit{XRView} interface, as it already contains the related \textit{projectionMatrix} attribute. 
To this end, we suggest extending the WebXR matrix notion to 9 element 3x3 matrices to accommodate the $\mathbf{K}$ matrix. Although it duplicates some information, it can be passed to computer vision algorithms without conversion --- similarly to how \textit{projectionMatrix} can be directly passed to WebGL.
Furthermore, an attribute storing $\mathbf{d}$ and the distortion model must be added.

The distortion model attribute should also be added to \textit{XRRenderState} for allowing applications to request a specific model as discussed in the section above --- similarly to how developers request a specific \textit{depthNear}.

This would enable browsers to transparently provide calibration data as provided by our service through the WebXR API. Alternatively browser vendors could opt to bundle a set of calibrations for popular cameras directly with the browser.

\section{Conclusion \& Future Work}
\label{sec:conclusion}

We have presented a calibration aggregation service, which allows the general deployment of web-based computer vision algorithms. Previously these would have been limited to systems where WebXR back-ends like ARKit or ARCore were available. The presented service also guides end-users through the task of calibration, enabling them to use cameras that were not considered by the developers of a particular computer vision algorithm. This property is beneficial for both users and developers of computer vision on the web.
At this we have evaluated the shortcomings of the current WebXR API draft end suggested extensions that can make the whole process transparent for the end-user.

However, additional support by the browsers might be needed to allow matching AR visualization. One possibility is to support image remapping through the WebXR API to allow rectification as shown in Figure \ref{fig:remap}. Alternatively the WebGL API could be extended to support the reverse direction, namely distorted rendering.
However, actual usage patterns should be analyzed to decide whether this would be beneficial or whether it is sufficient to offload these tasks to client libraries like OpenCV.js.

Furthermore, it needs to be evaluated whether our calibration key is sufficient to identify the various cameras and devices or if we have to use more sophisticated fingerprinting.

\bibliographystyle{ACM-Reference-Format}
\bibliography{bibliography} 

\end{document}